\documentclass[twoside]{article}

\usepackage[accepted]{aistats2021}
\usepackage[round]{natbib}

\usepackage[utf8]{inputenc} 
\usepackage[T1]{fontenc}    
\usepackage{hyperref}       
\usepackage{url}            
\usepackage{booktabs}       
\usepackage{amsfonts}       
\usepackage{nicefrac}       
\usepackage{microtype}      

\usepackage{graphicx}
\usepackage{xcolor}
\usepackage{colortbl}
\usepackage{array}
\usepackage{pifont}
\usepackage{multirow}
\usepackage{subfig}
\usepackage{amsmath}
\usepackage{amsthm}
\usepackage{amssymb}%
\usepackage{bbm}
\usepackage{makecell}
\usepackage{hhline}
\usepackage{commath}
\usepackage{ulem}
\usepackage[font=small,labelfont=bf]{caption}
\usepackage{wrapfig}
\usepackage{breakcites}
\usepackage{sidecap}

\DeclareMathOperator{\EE}{\mathbb{E}} %
\DeclareMathOperator{\Z}{\mathbf{Z}} %
\DeclareMathOperator{\C}{\mathbf{C}} %
\DeclareMathOperator{\W}{\mathbf{W}} %
\DeclareMathOperator{\bb}{\mathbf{b}} %
\DeclareMathOperator{\x}{\mathbf{x}} %
\DeclareMathOperator{\z}{\mathbf{z}} %
\DeclareMathOperator{\y}{\mathbf{y}} %
\DeclareMathOperator{\nn}{\mathcal{N}_{\mathcal{A}}} %
\begin{document}

%

%

\twocolumn[

\aistatstitle{Statistical Characteristics of Deep Representations: \\ An Empirical Investigation}

\aistatsauthor{ Daeyoung Choi\textsuperscript{*}, Kyungeun Lee\textsuperscript{*}, Duhun Hwang, and Wonjong Rhee}

\aistatsaddress{ Department of Transdisciplinary Studies \\
Seoul National University \\
Seoul, 08826, South Korea \\
\{choid, ruddms0415, yelobean, wrhee\}@snu.ac.kr} ]

\begin{abstract}
  In this study, the effects of eight representation regularization methods are investigated, including two newly developed rank regularizers (RR). 
  The investigation shows that the statistical characteristics of representations such as correlation, sparsity, and rank can be manipulated as intended, during training. Furthermore, it is possible to improve the baseline performance simply by trying all the representation regularizers and fine-tuning the strength of their effects. 
  In contrast to performance improvement, no consistent relationship between performance and statistical characteristics was observable.
  The results indicate that manipulation of statistical characteristics can be helpful for improving performance, but only indirectly through its influence on learning dynamics or its tuning effects.

\end{abstract}

\section{Introduction}
\label{sec:intro}

\begin{figure}[t]
  \includegraphics[width=7.8cm]{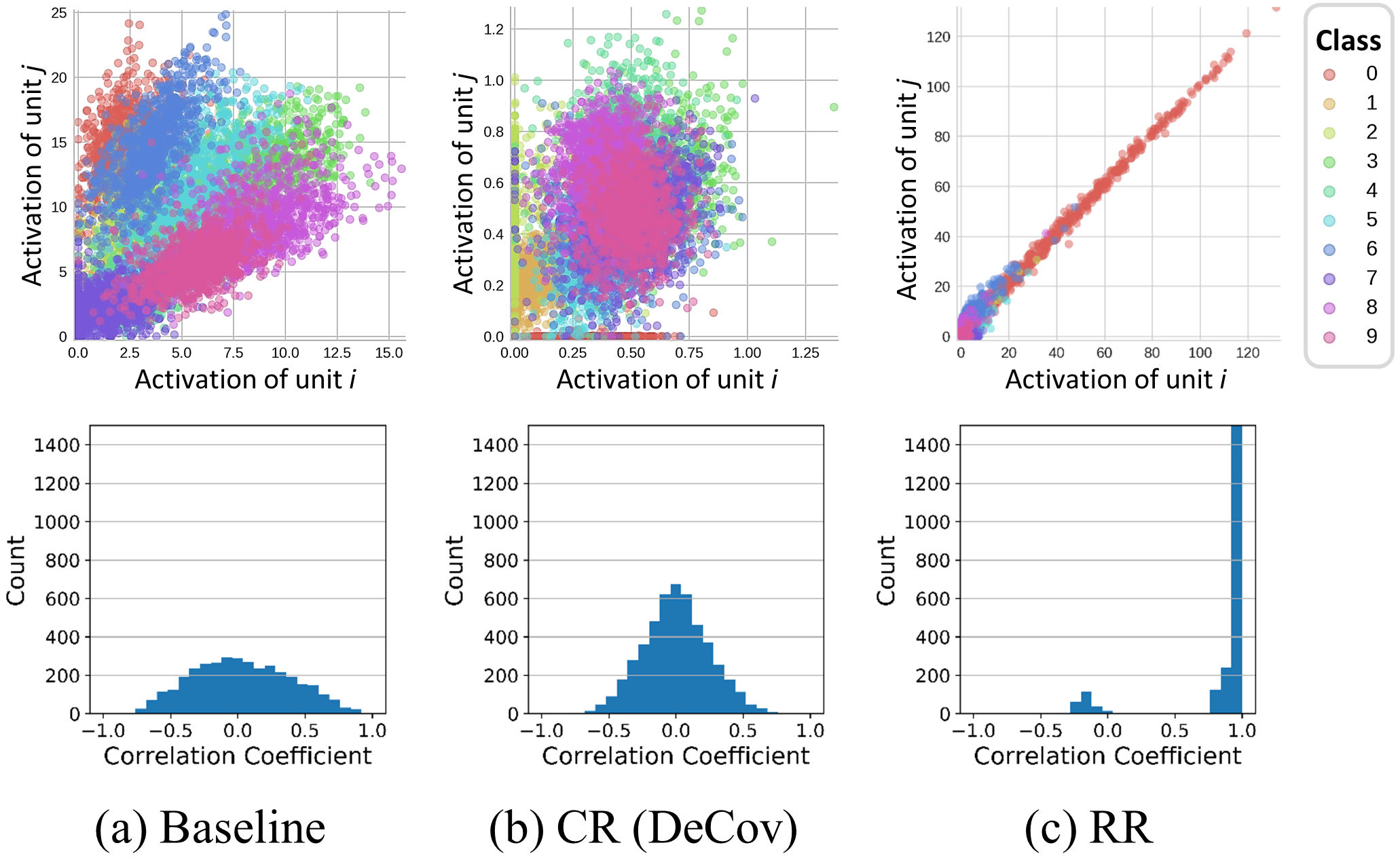}
  \caption{Two representative units' activation scatter plots (upper plots) and histograms of the correlation coefficient distribution (lower plots) for the MNIST dataset. For a 6-layer MLP  with 100 units for each layer, the fifth layer's activation vectors, calculated using 10,000 test samples, were used to select two neurons randomly and to generate the plots. (a) The baseline model shows a moderate correlation. (b) CR (DeCov) shows very low correlation. (c) Rank regularizer (RR) has completely different characteristics (high correlation) compared to CR. Despite exhibiting totally different representation characteristics, the performances of the three models are comparable.}
  \label{fig:correlation_example}
\end{figure}

A learned representation can affect the performance of deep neural networks; the distributed and deep natures of the representation are the essential elements for the success of deep learning \citep{bengio2013representation}. Owing to the depth, deep networks have a greater \textit{expressiveness} compared to other machine learning algorithms ~\citep{hinton1986learning}, or shallow networks 
~\citep{montufar2014number,telgarsky2015representation,eldan2016power,raghu2016expressive}. In addition to the distributed and deep natures that have been intensively studied, a hidden layer's \textit{representation characteristics} are considered to be important as well. Nonetheless, a relatively limited number of studies have been conducted on this matter. The goal of the present study is to better understand representation characteristics. In this study, the meaning of representation is restricted to \textit{the activation vector of a single hidden layer}, and representation characteristics refer to \textit{the statistical characteristics of the activation vector, such as correlation and sparsity}.

In the past several years, dropout~\citep{srivastava2014dropout} and batch normalization (BN)~\citep{ioffe2015batch} have become essential regularization options, in addition to the default options of L1 \citep{hoerl1970ridge} and L2 \citep{tibshirani1996regression} weight regularizations. 
Additionally, manipulating representation characteristics has become increasingly popular for the improvement of performance~\citep{glorot2011deep,cogswell2015reducing,xiong2016regularizing,liao2016learning,wen2016discriminative,belharbi2017neural,choi2018utilizing,hofer2020topologically}. 
The regularization methods often lead to improved performance, but rigorous explanation has been missing. Instead, it has been implied or conjectured that the manipulation of the representation can lead to improved performance because of known and relevant concepts in machine learning (typically, a reduced generalization gap is quoted as the supporting empirical evidence). For instance, reduced co-adaptation (that is closely related to the correlation of the representation) has been put forth as a possible reason for the good performance of dropout; sparser or less correlated representations have been argued as better representations because the number of true underlying features must be limited. As another example, reducing covariate shift was the reason for inventing batch normalization, but later it was shown that no reduction of internal covariate shift is observable \citep{santurkar2018does}. Instead,~\cite{santurkar2018does} show that BN makes the optimization landscape significantly smoother,~\cite{bjorck2018understanding} demonstrate how large gradient updates can result in diverging loss and activations growing uncontrollably with network depth and how BN avoids these, and~\cite{de2020batch} provide a downscaling method that can replace BN at a comparable performance. As a more general result on regularizing  representation,~\cite{locatello2019challenging} recently showed that aiming for certain activation characteristics (`disentangled representations' in their work, in contrast to `representation characteristics' in our work) is not as universally meaningful as is often assumed in unsupervised learning. 

In our study, statistical characteristics of deep representations are investigated for common supervised learning tasks. As a basic framework for the study, an extensive set of representation regularizers are considered, including a baseline model (no regularization). A total of eight regularization options are investigated to examine six characteristics of deep representations. Among the eight, rank regularizer (RR) and class-wise rank regularizer (cw-RR) were newly designed and tested in this study because of their association with important representation characteristics, such as correlation and rank. The regularizers aim to decrease the rank of representation (increase the correlation of representation) by reducing the stable rank of each mini-batch activation from the all-class samples (RR) or the same-class samples (cw-RR). 

Some examples of the representations found with the regularizers are shown in Figure \ref{fig:correlation_example} (more examples are shown in Figure \ref{fig:scatter}). In the figure, correlation characteristics vary largely, depending on which regularizers are used. RR shows a strong correlation and has a performance comparable to CR (DeCov), as shown in the lower plot of Figure \ref{fig:correlation_example}. The comparable performance of RR under strong correlation is precisely the opposite of what has been conjectured for DeCov \citep{cogswell2015reducing}. 

As we will show in Section \ref{sec:empirical_results}, representation characteristics can be manipulated as intended, by applying a variety of regularizers. Additionally, all these regularizers, as a set, can be a useful tool for improving performance. However, the problem is that there is (perhaps unsurprisingly) no distinct pattern that can be used to assess which regularizer (representation characteristic) is likely to be helpful for a given task.  All that can be concluded is that some regularizers would be helpful for any given task; but it is not possible to find which ones would qualify to be helpful for the same. In this paper, we do not claim that deep learning practitioners should not attempt to change representation characteristics owing to this problem. Instead, we empirically show that representation regularization can be a useful option for improving performance at the cost of tedious tuning. Despite of inconspicuous relationship between representation characteristics and performance, representation regularization can work as proxies that affect the learning dynamics of deep network training and thus indirectly improve the performance.  

\section{Related Works}
\label{sec:related_work}

\begin{table*}[!t]
\centering
\caption{Symbols and expressions of representation characteristics.}
\label{table:characteristics}
\renewcommand{\arraystretch}{1.2}
\resizebox{\textwidth}{!}{%
\setlength\tabcolsep{2.5pt}
\begin{tabular}{ccl} 
\toprule 
Characteristic  & Symbol & \multicolumn{1}{c}{Expression}      \\ \hline
\textsc{Amplitude}   & $\bar{|z|}$ 
& $\EE_i [|\z_{l,i}|]$  
             \\ \hline
\textsc{Covariance}     & $\bar c$ 
& $\EE_{i\neq j} [|c_{i,j}|]$, 
where $c_{i,j} \triangleq \{\C_l\}_{i,j} = \EE [(\z_{l,i} - \mu_{z_{l,i}})(\z_{l,j} - \mu_{z_{l,j}})]$
     \\ \hline
\textsc{Correlation}    & $\bar \rho$    
& $\EE_{i\neq j} [|\rho_{i,j}|]$, 
where $\rho_{i,j} \triangleq \{\C_l\}_{i,j} /\sigma_{z_{l,i}}\sigma_{z_{l,j}}
=\EE [(\z_{l,i} - \mu_{z_{l,i}})(\z_{l,j} - \mu_{z_{l,j}})]/\sigma_{z_{l,i}}\sigma_{z_{l,j}}$ 
     \\ \hline
\textsc{Sparsity}       & $P_s$       
& $\EE_{i,n} [\mathbbm{1}{(z_{l,i}^{n})}]$, 
where $\mathbbm{1}$ is an indicator function whose output is 1 only when $z_{l,i}^{n}=0$
          \\ \hline
\textsc{Dead unit}       & $P_d$       
& $\EE_{i} [\mathbbm{1}{(z_{l,i})}]$, 
where $\mathbbm{1}$ is an indicator function whose output is 1 only when $z_{l,i}^n=0$ for all $n=1,..,N$
          \\ \hline
\textsc{Rank}      & $r$      
& $rank(\C_l)$; numerical evaluations are approximated as the stable rank $\norm{\mathbf{C}_{l}}_{F}^{2}/\norm{\mathbf{C}_{l}}_{2}^{2}$
             \\ 
\bottomrule
\end{tabular}
}
\end{table*}

\textbf{Popular Regularization Methods} \quad
Many distinct regularization methods have been developed for deep learning. The most traditional methods are L1 and L2 weight regularizations (L1W, L2W). Dropout \citep{srivastava2014dropout} and batch normalization \citep{ioffe2015batch} have shown large performance improvements in the context of many interesting tasks. With the extended definition of regularization in \citep{goodfellow2016deep}, many other methods such as data augmentation, adversarial training, and multi-task learning can be considered as regularization methods too. In this work, however, we limit our focus to the traditional, dropout, batch normalization, and representation regularizations.

\noindent\textbf{Representation Regularization Methods} \quad
A representation regularizer explicitly aims to modify a statistical property of the activation vectors, typically by using a penalty. One of the earliest representation regularizers is the L1 representation regularizer \citep{glorot2011deep}, and it applies an L1 penalty to the activation vector instead of the weight vectors. It encourages sparsity in the representation, and it is called L1R in this work. \cite{cheung2014discovering} reduce a sum-squared cross-covariance between autoencoding and label unit activations to disentangle representations. Similarly, \cite{cogswell2015reducing} suggest DeCov that utilizes a penalizing loss function to reduce activation covariance among hidden units. \cite{choi2018utilizing} consider the extension to class-wise regularization and provide four representation regularizers: CR (Covariance regularizer), cw-CR (class-wise covariance regularizer), VR (variance regularizer), and cw-VR (class-wise variance regularizer). Among them, CR is equivalent to DeCov. 

\noindent\textbf{Role of Explicit Regularizations} \quad
\cite{zhang2016understanding} showed that explicit regularizations such as L2W  and dropout are not directly responsible for reducing or controlling the generalization error. Rather, they argue that performance improvement can be because of a tuning effect. \cite{arpit2017closer} investigated the impact of explicit regularization on the memorization speed and generalization. 

\noindent\textbf{Generalization of Deep Networks} \quad
Recently, generalization bounds for deep neural networks have been heavily studied\citep{neyshabur2015norm, dziugaite2017computing, bartlett2017spectrally, arora2018stronger, jiang2019fantastic}. Complexity measure is a core component of the generalization bounds. For instance, \cite{neyshabur2015norm, sanyal2019stable} showed that norm-based regularizations can control network complexity. Most of the bounds found so far, however, are far from being tight for deep networks.

\section{Representation Characteristics and Regularizers}
\label{sec:characteristics}


\subsection{Representation Characteristics}

Consider a neural network $\nn$
whose architecture $\mathcal{A}$ is fixed and 
the weights for the $l$\textsuperscript{th} layer are given by $\{\W_l\}$ and $\{\bb_l\}$ after training.
We write $\nn=(\W, \bb)$ to 
denote a network and $\y$ or $\nn(\x)$ to refer to its deterministic output for a given input $\x$. The index $l$ is omitted when the meaning is obvious.
The activation vector of the $l$\textsuperscript{th} layer for the given input $\x$ is denoted as $\z_l(\x)$ or simply $\z_l$, 
and the $i$\textsuperscript{th} element of $\z_l$ is denoted as $z_{l,i}$. The mean, variance, and standard deviation of $z_{l,i}$ over $p(\x)$ are 
defined as $\mu_{z_{l,i}}$, $v_{z_{l,i}}$, and $\sigma_{z_{l,i}}$, respectively.
The covariance of $\z_l$ is defined as $\C_l$. 
The definitions of class-wise statistics are included in Section A of the supplementary materials.

The basic representation characteristics can be summarized as in Table \ref{table:characteristics}. Since the true distribution of the data is not accessible, the numerical results in the following sections are evaluated using the empirical distribution of the test dataset. For instance, $\C_l$ is calculated as the covariance matrix of $N$ activation vectors $\{\z_l^1, ... , \z_l^N\}$ where $\z_l^n$ corresponds to the activation vector for the $n$\textsuperscript{th} test data example, $\x^n$. Rank can be calculated by examining $\C_l$, but often there are small eigenvalues that hinder a proper assessment of the rank. Therefore, \textit{stable rank} is evaluated instead. 

\subsection{Representation Regularizers}
In this study, mainly eight options are considered: the baseline model (no regularizer) and seven models of representation regularizers (CR, cw-CR, VR, cw-VR, L1R, RR, cw-RR). Even though dropout and BN do not explicitly target to modify representation characteristics, they were also studied together because the popular regularizers certainly affect the representation characteristics. 
Regularization terms are added to the original cost function as penalty regularizers. The total cost function $\widetilde{J}$ can be denoted as
\begin{align}
    \widetilde{J}=J+\lambda{\Omega}(\mathbf{z}),
\end{align}
where $\lambda$ is the loss weight ($\lambda \in [0, \infty)$).
Each regularizer targets a different statistical characteristic of the representations. For example, CR and VR reduce covariance and variance of the activations calculated from all-class samples, respectively. L1R decreases the absolute amplitude of activations calculated from all-class samples to make the activations sparser. Regularizers with prefix `cw-' are the class-wise counterparts of all-class regularizers. All the loss functions are summarized in Section A of the supplementary material.

\textbf{Rank Regularizer}  \quad
In deep learning, a low-rank approximation of convolutional filters \citep{jaderberg2014speeding,lebedev2014speeding,tai2015convolutional} and weight matrices \citep{nakkiran2015compressing,masana2017domain,alvarez2017compression} has been widely used for network compression and fast network training. The recent work by \cite{sanyal2019stable} proposes stable rank normalization (SRN), which can improve the generalization of the network in  classification tasks. Given the available literature, regularization methods are typically applied to weights, and not to activations. However, in this study, RR and cw-RR are applied to activations as penalty regularizers.  RR is designed to encourage a lower rank of representations and is used while training the network. Because the usual definition of rank can be very sensitive to small  singular values, we use \textit{stable rank} of the activation matrix $\Z=[\z_{l}^{1},\dots ,\z_{l}^{N_{MB}}]^{T}$ as a surrogate. Note that $N_{MB}$ instead of $N$ activation vectors are used for each mini-batch. The stable rank of $\Z$ is defined as 
\begin{align}
    \Omega_{RR}=\frac{\norm{\Z}_{F}^{2}}{\norm{\Z}_{2}^{2}}  =\frac{\sum_i s_i^2} {\max_i s_i^2},
\end{align}
where $\norm{\Z}_{F}$ is the Frobenius norm, $\norm{\Z}_{2}$ is the spectral norm, and $\{s_i\}$ are the singular values of $\Z$. From $\frac{\sum_i s_i^2} {\max_i s_i^2}$, it can be clearly seen that stable rank is upper-bounded by the rank that counts strictly positive singular values. As the spectral norm is based on singular value decomposition, calculating the derivative of the stable rank for every mini-batch 
is a computationally heavy operation. To reduce the computational burden, we apply an approximation using a special case of H{\"o}lder's inequality.
\begin{align}
    \Omega_{RR}=
    \frac{\norm{\Z}_{F}^{2}}{\norm{\Z}_{2}^{2}}
    & =\frac{\text{trace}(\Z^{T}\Z)}{\norm{\Z}_{2}^{2}} \\
    & \geq \frac{\text{trace}(\Z^{T}\Z)}{\norm{\Z}_{1}\norm{\Z}_{\infty}} \\
    & =\frac{\sum_{i,n}(z_{i}^{n})^{2}}
       {(\max_{i}\sum_{n=1}^{N_{MB}}{|z_{i}^{n}|})(\max_{n}\sum_{i=1}^{M}{|z_{i}^{n}|})} 
\end{align}
The inequality $\norm{\Z}_{2}\leq \sqrt{\norm{\Z}_{1}\norm{\Z}_{\infty}}$ is used, where $\norm{\Z}_{1}$ is the maximum absolute column-wise sum of the matrix $\Z$ (sum of all activation values of unit $i$) and $\norm{\Z}_{\infty}$ is the maximum absolute row-wise sum of the matrix $\Z$ (sum of all activation values of sample $n$). The extension of RR to cw-RR is straightforward.


\section{Experiments}
\label{sec:empirical_results}

In this section, it is empirically shown that the regularization affects the statistical characteristics of deep representations. The relationship between performance and the representation characteristics is also examined. Finally,  performance results on a variety of tasks are presented.

\subsection{Experimental Settings}

As examples of simple networks, we used a 6-layer MLP for the MNIST dataset, and a CNN with four convolutional layers and one fully-connected layer for the CIFAR-10/100 dataset. (In this paper, we call them `MLP' and `CNN' respectively, for convenience.) As examples of sophisticated networks, VGG-16 on the CIFAR-10/100, ResNet-18/50 on the ImageNet/Tiny-ImageNet datasets were used. For ResNet, a single fully-connected layer was added following the last average pooling layer. Validation performance was evaluated with different loss weights \{0.001, 0.01, 0.1, 1, 10, 100, 1000\}, and the one with the best validation performance for each regularizer and condition was chosen for testing. For ResNet, pre-trained models were fine-tuned with the regularizers. Each training trial was repeated five times unless mentioned; the mean and standard deviation of the five trials are reported. The mini-batch size was set to 100 for MLP (MNIST) and CNN (CIFAR-10), 128 for VGG-16 (CIFAR-10) and ResNet-50 (Tiny-ImageNet), and 256 for ResNet-18 (ImageNet). For CIFAR-100 that has 100 classes, mini-batch size of 500 was used to calculate meaningful class-wise statistics. Experiments with class-wise regularizers were not performed for ImageNet and Tiny-Imagenet datasets to avoid inefficient training of large mini-batch size. More details of the experimental settings can be found in Section B of the supplementary material.

\subsection{Effect of Regularization on Representation Characteristics}

\begin{table*}[!t]
\begin{minipage}[b]{1.\linewidth}
\centering
\includegraphics[width=0.98\textwidth]{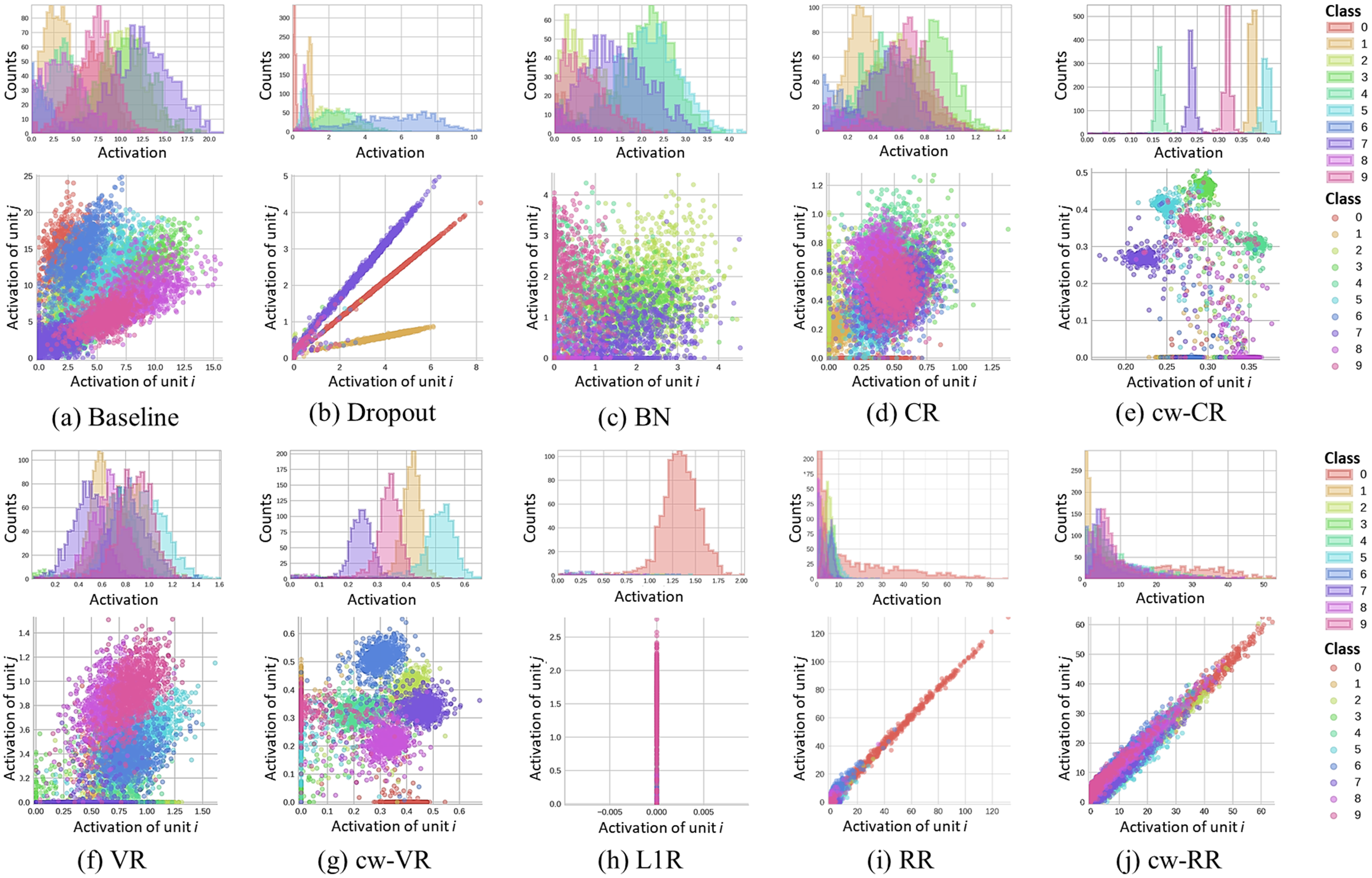}
\captionof{figure}{Activation histogram of a single unit (upper plots) and the activation scatter plots of two randomly chosen units (lower plots) for a 6-layer MLP trained with the MNIST dataset. The plots were produced in the same way as in Figure \ref{fig:correlation_example}.
\textbf{(upper)} The baseline has a large class-wise variance and inter-class overlaps; BN and CR (covariance regularizer) show similar properties. Dropout looks completely different where activation values are more spread out. cw-CR and cw-VR show well-separated activation distributions because they are regularized class-wise. L1R increases the sparsity of representation.
\textbf{(lower)} As mentioned in the caption of Figure \ref{fig:correlation_example}, CR, RR, and cw-RR show completely different patterns. cw-CR and cw-VR show low correlation per class because they are regularized class-wise.} 
\label{fig:scatter}
\end{minipage}
\\

\begin{minipage}[b]{1.0\linewidth}
\centering
\caption{Statistical characteristics of learned representations. The characteristics of MLP were generated in the same way as in Figure \ref{fig:correlation_example}. For ResNet, one fully-connected layer was added next to the last average pooling layer and regularizers were applied on it. One can observe that the characteristics are modified, as initially predicted (indicated in \textbf{bold}).} 
\label{table:stat_property}
\resizebox{\textwidth}{!}{%
\setlength\tabcolsep{3pt} 
\centering
\begin{small}
\resizebox{\textwidth}{!}{%
\begin{tabular}{llrrrrrrrr}
\toprule
Data \& Net.& Reg.   & Accuracy (\%) &  \textsc{Amplitude}   & \textsc{Covariance}  & \textsc{Correlation} & \textsc{Sparsity}  & \textsc{Dead unit}     & \textsc{Rank} \\ \hline
\multirow{4}{*}{\rotatebox[origin=c]{90}{\makecell{MNIST \\ \\ MLP}}}
& Baseline      & 97.15 &   4.93   & 2.08  & 0.27  & 0.34  & 0.13  & 2.41 \\  
& CR            & 97.50 &   0.50   & \textbf{0.01} & \textbf{0.19} & 0.40 & 0.03 & 7.12  \\ 
& L1R           & 97.65 &   1.29   & 0.03  & 0.40  & \textbf{0.97} & \textbf{0.39} & 5.94  \\ 
& RR            & 97.19 &   7.23   & 226.20 & 0.90  & 0.43  & 0.18  & \textbf{1.00} \\ \hline   
\multirow{4}{*}{\rotatebox[origin=c]{90}{\makecell{Tiny- \\ ImageNet \\ ResNet-50}}} 
& Baseline      & 78.56 &   1.06   & 0.155  & 0.08  & 0.436  & 0.00  & 6.51 \\ 
& CR            & 78.14 &   0.26   & \textbf{0.007} & \textbf{0.04} & 0.585 & 0.00 & 26.09  \\  
& L1R           & 78.32 &   0.22   & 0.016  & 0.05  & \textbf{0.780} & 0.00 & 5.36  \\ 
& RR            & 77.99 &   1.59   & 0.204 & 0.12 & 0.155 & 0.00 & \textbf{1.46} \\
\hline  
\multirow{4}{*}{\rotatebox[origin=c]{90}{\makecell{ImageNet \\ \\ ResNet-18}}} 
& Baseline      & 70.34 &   0.90   & 0.049  & 0.062  & 0.010  & 0.00  & 6.46 \\ 
& CR            & 68.76 &   0.52   & \textbf{0.005} & \textbf{0.051} & 0.000 & 0.00 & 22.46  \\  
& L1R           & 69.51 &   0.83   & 0.067  & 0.078  & \textbf{0.010} & 0.00 & 2.40  \\ 
& RR            & 69.75 &   0.92   & 20.448 & 0.968 & 0.012 & 0.00 & \textbf{1.00} \\
\bottomrule
\end{tabular}
}
\end{small}
}
\end{minipage}
\end{table*}

The representation characteristics were visually and quantitatively investigated, as shown in Figure \ref{fig:scatter} and Table \ref{table:stat_property}. 
In Figure \ref{fig:scatter}, it can be observed that the representation characteristics exhibit large variations depending on the choice of the regularizer. In particular, dropout shows a strong pair-wise correlation, as shown in the lower plot of Figure \ref{fig:scatter}(b). This is precisely the opposite of what has been believed for dropout.
Even though not shown, the visualization of the CIFAR-10/100, the Tiny-ImageNet, and the ImageNet datasets showed similar patterns as in Figure \ref{fig:scatter} (the patterns were less distinct for the class-wise regularizers). The plots of the top three principal components of the representations are included in Section C of the supplementary material, to present distinct global trends of the representations.

Our quantitative result confirms the visualization. Each characteristic was obtained by applying the largest loss weight possible while maintaining comparable performance with the baseline model. The result confirms that the statistical characteristics targeted by each regularizer are manipulated as expected \textbf{(Bold)} in Table \ref{table:stat_property}. In particular, RR regularizes the stable rank, and thus works as intended. RR (highly correlated representations) shows comparable performance to CR (decorrelated representations). This result is somewhat counter-intuitive to the conventional wisdom. 

\begin{figure*}[!t]
\centering
\subfloat[\small{Correlation} (MNIST)]{
    \includegraphics[width=0.24\textwidth]{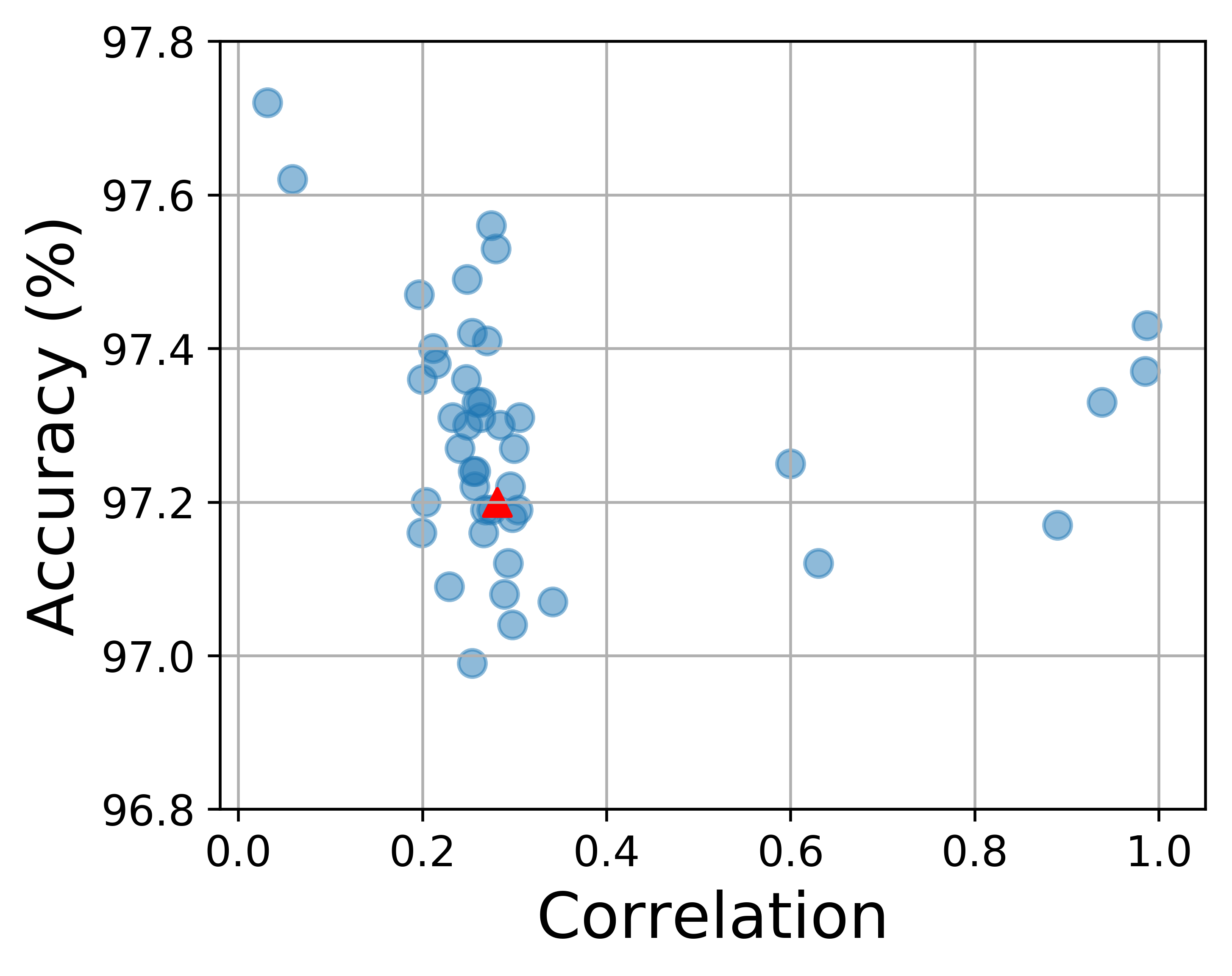}
}
\subfloat[\small{Correlation} (CIFAR-100)]{
    \includegraphics[width=0.23\textwidth]{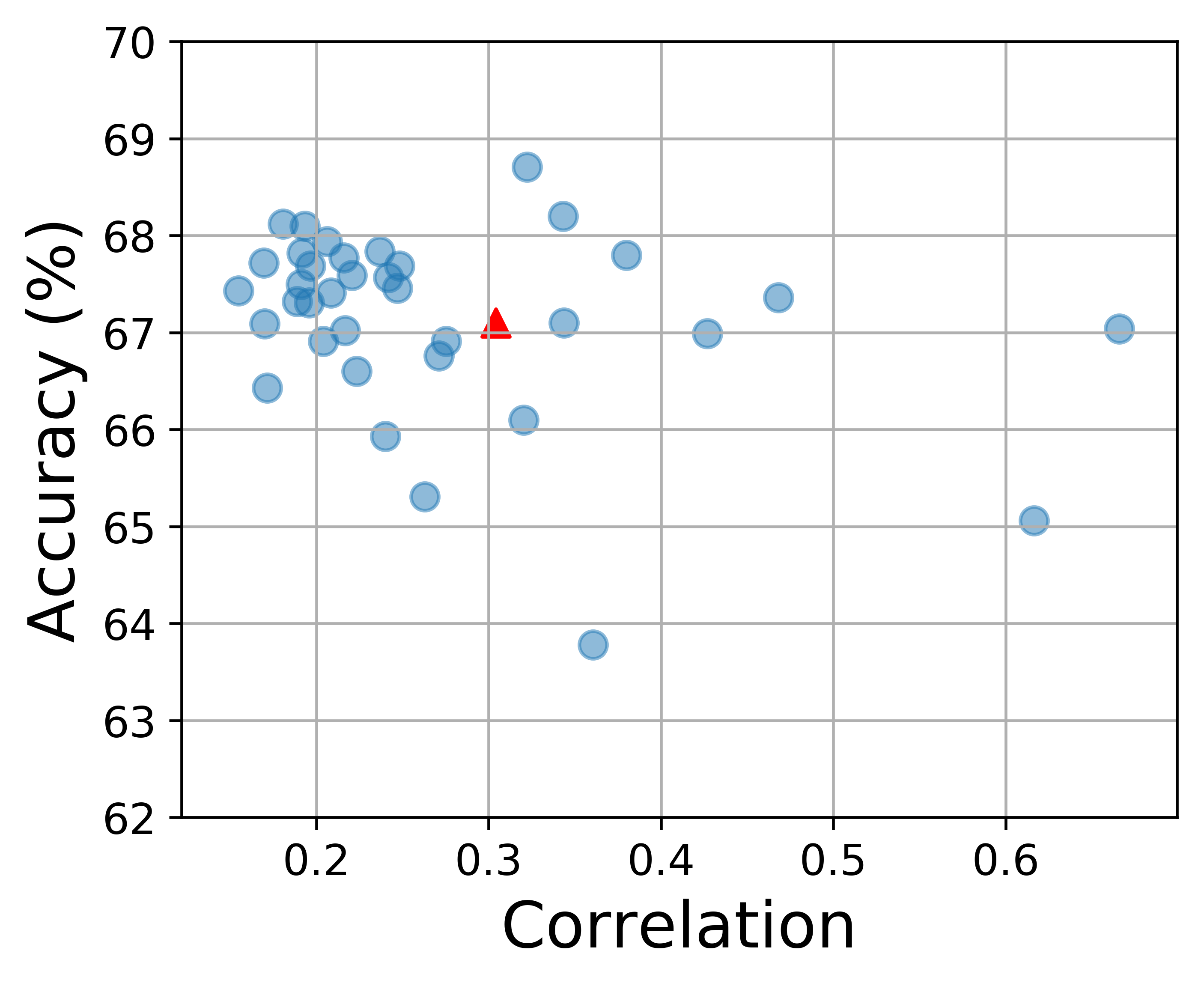}
}
\subfloat[\small{Sparsity} (MNIST)]{
    \includegraphics[width=0.24\textwidth]{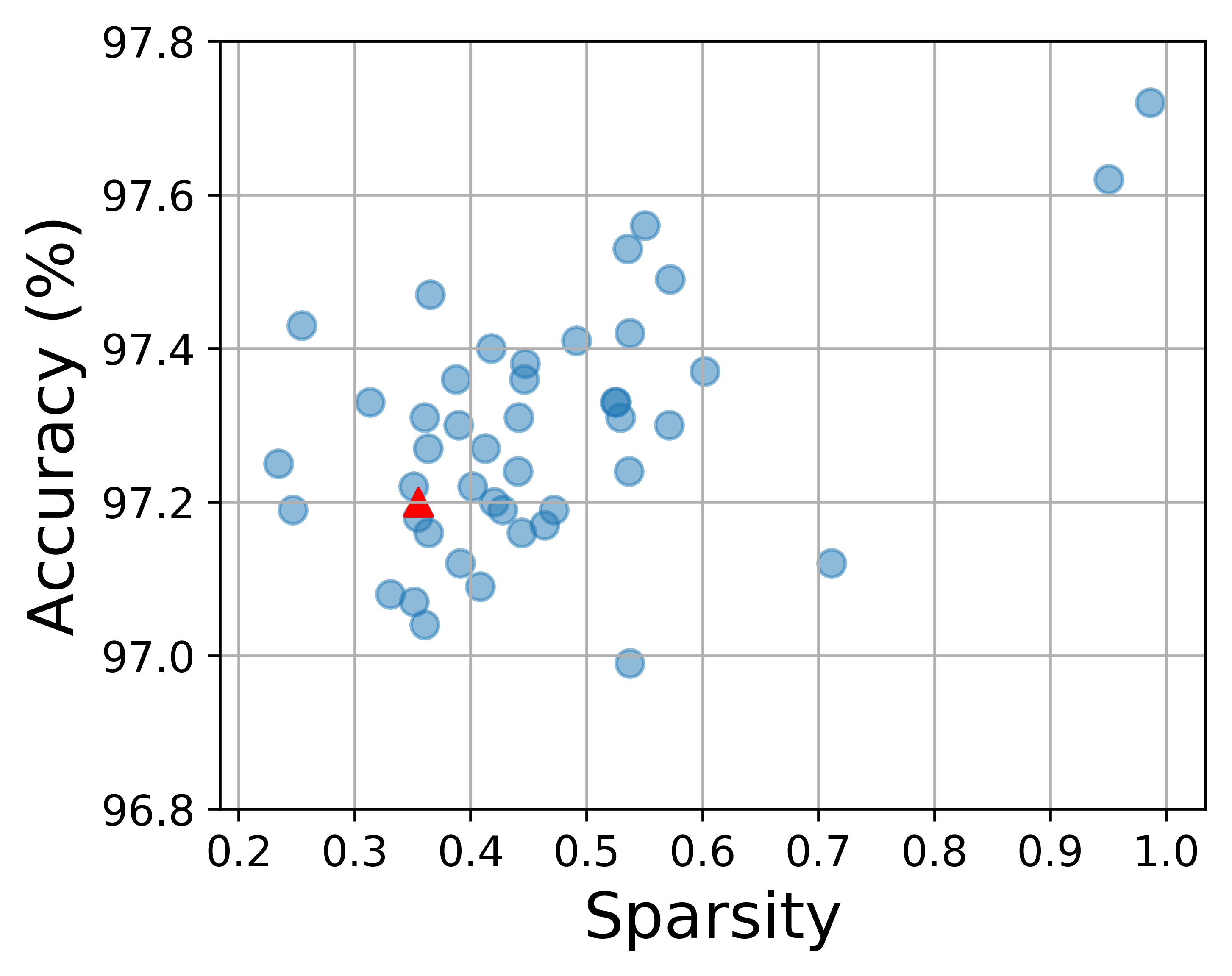}
}
\subfloat[\small{Sparsity} (CIFAR-100)]{
    \includegraphics[width=0.23\textwidth]{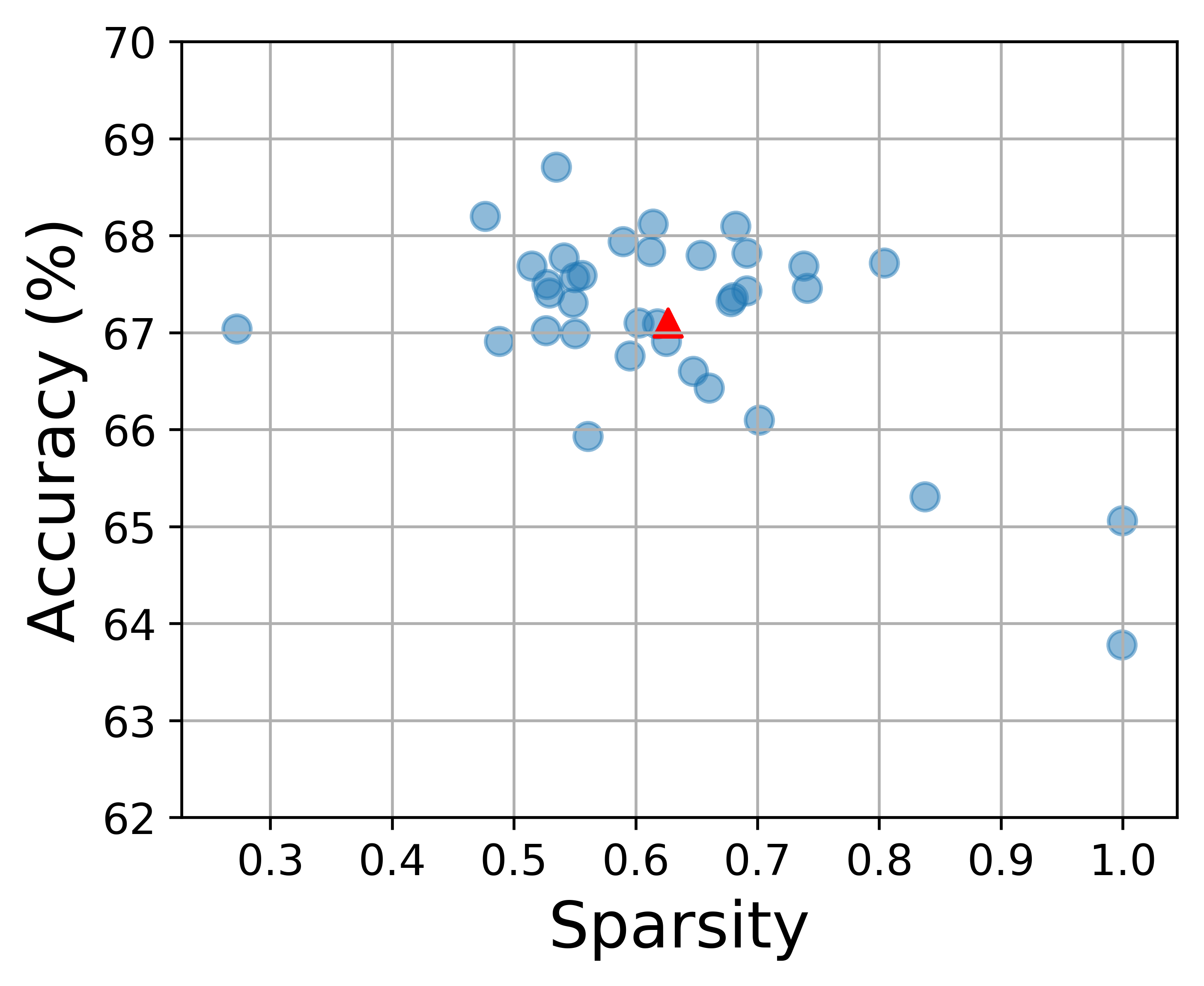}
}
\caption{Relationship between the representation characteristics and the performance on the MNIST (MLP) and the CIFAR-100 (VGG-16). Each blue point indicates a single pair of representation characteristic and performance, from the corresponding model that utilizes specific regularizer and loss weight. The red triangle indicates the baseline model. 
}
\label{fig:relation_perf_stats}
\bigskip
\centering
\includegraphics[width=0.90\textwidth]{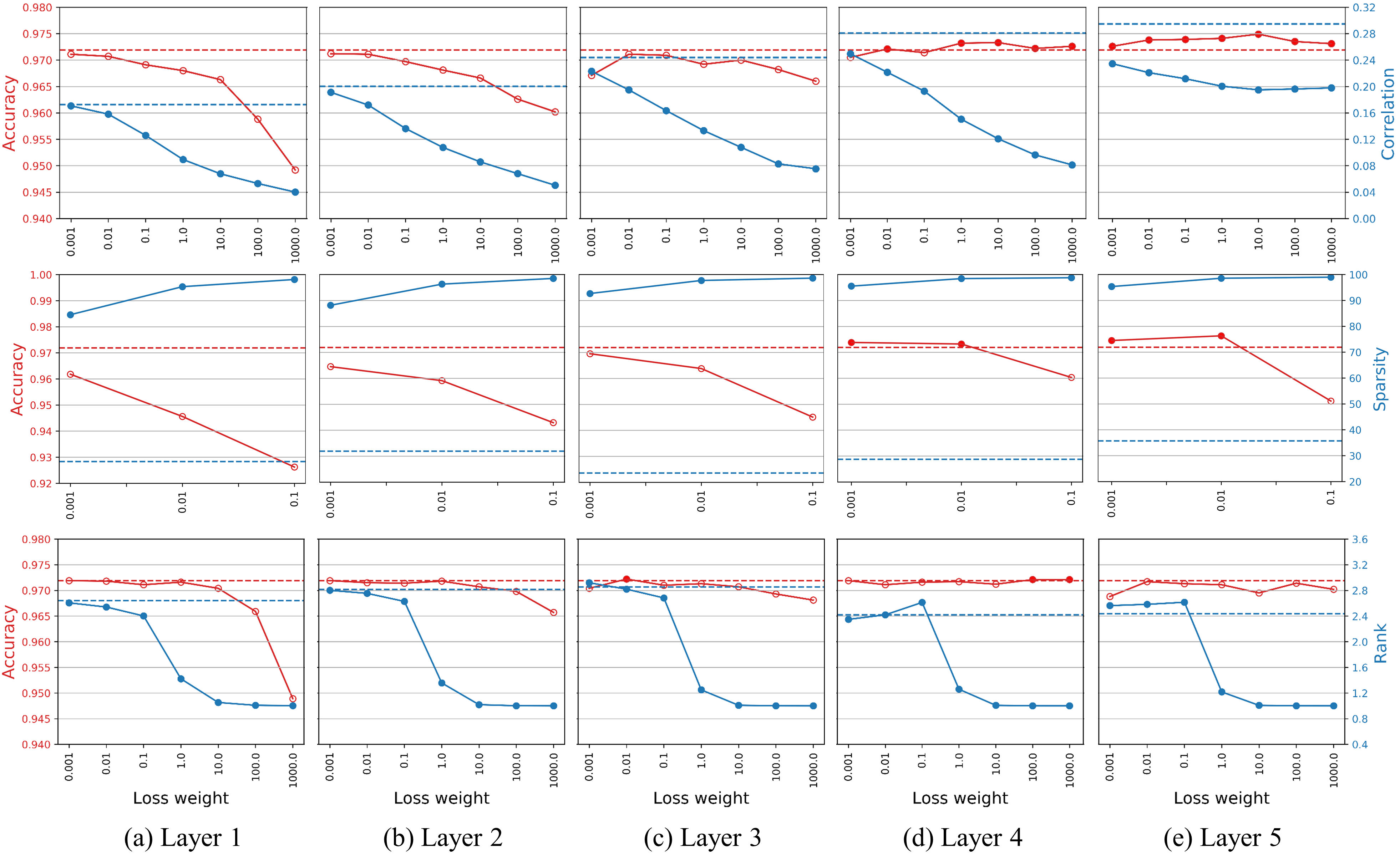}
\vskip 0.1in
\captionof{figure}{
Layer dependence of representation regularizations. Each plot was generated with the MLP on the MNIST in the same manner, as shown in Figure \ref{fig:correlation_example}. Regularizers were applied to all the layers. The top, middle, and bottom rows correspond to results of CR, L1R, and RR; the red and blue dotted lines indicate the baseline model's performance and the characteristics of each of its layers, respectively.  
Note that some models of L1R are excluded because they cannot be trained with loss weights that are greater than 0.1. } 
\label{fig:layer_dependency}
\end{figure*}

\begin{figure*}[!t]
\captionsetup[subfigure]{justification=centering}
\centering
\subfloat[text fot LoF][Data size

                        (MLP on MNIST)]{   
    \includegraphics[width=0.24\textwidth]{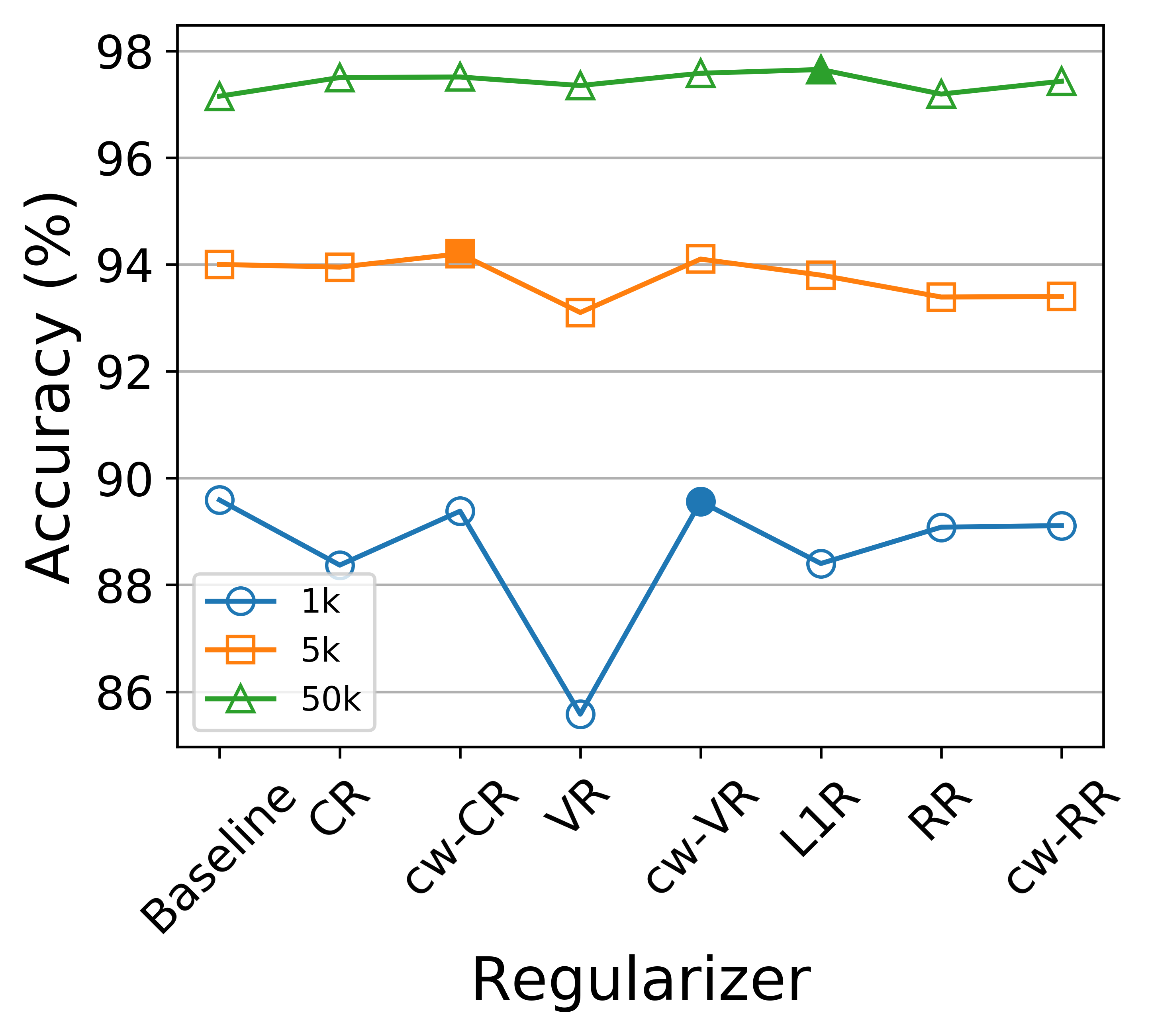} 
}
\subfloat[text fot LoF][Layer width 

                        (CNN on CIFAR-10)]{
    \includegraphics[width=0.24\textwidth]{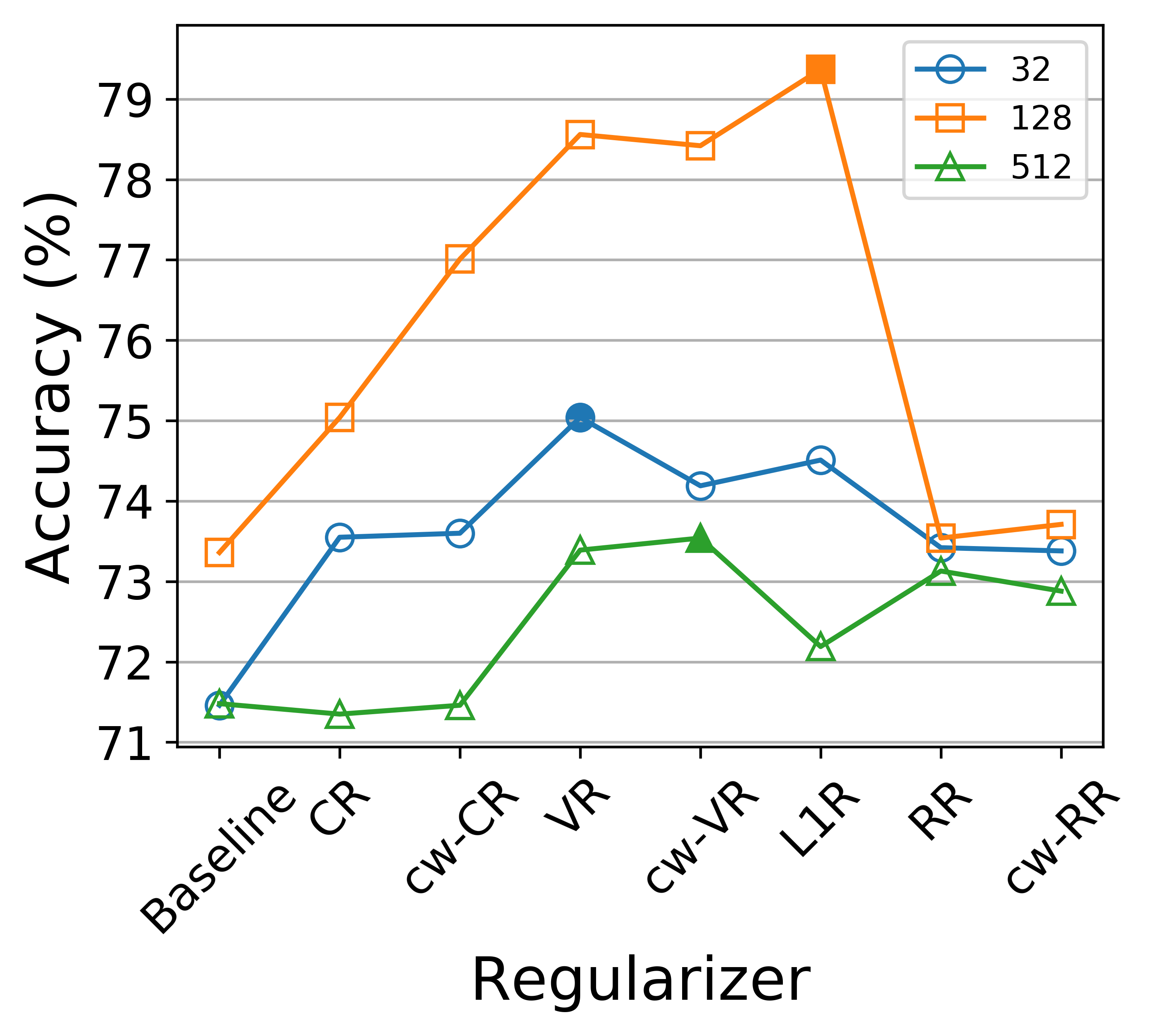}
}
\subfloat[text fot LoF][Optimizer 
                        
                        (CNN on CIFAR-10)]{
    \includegraphics[width=0.24\textwidth]{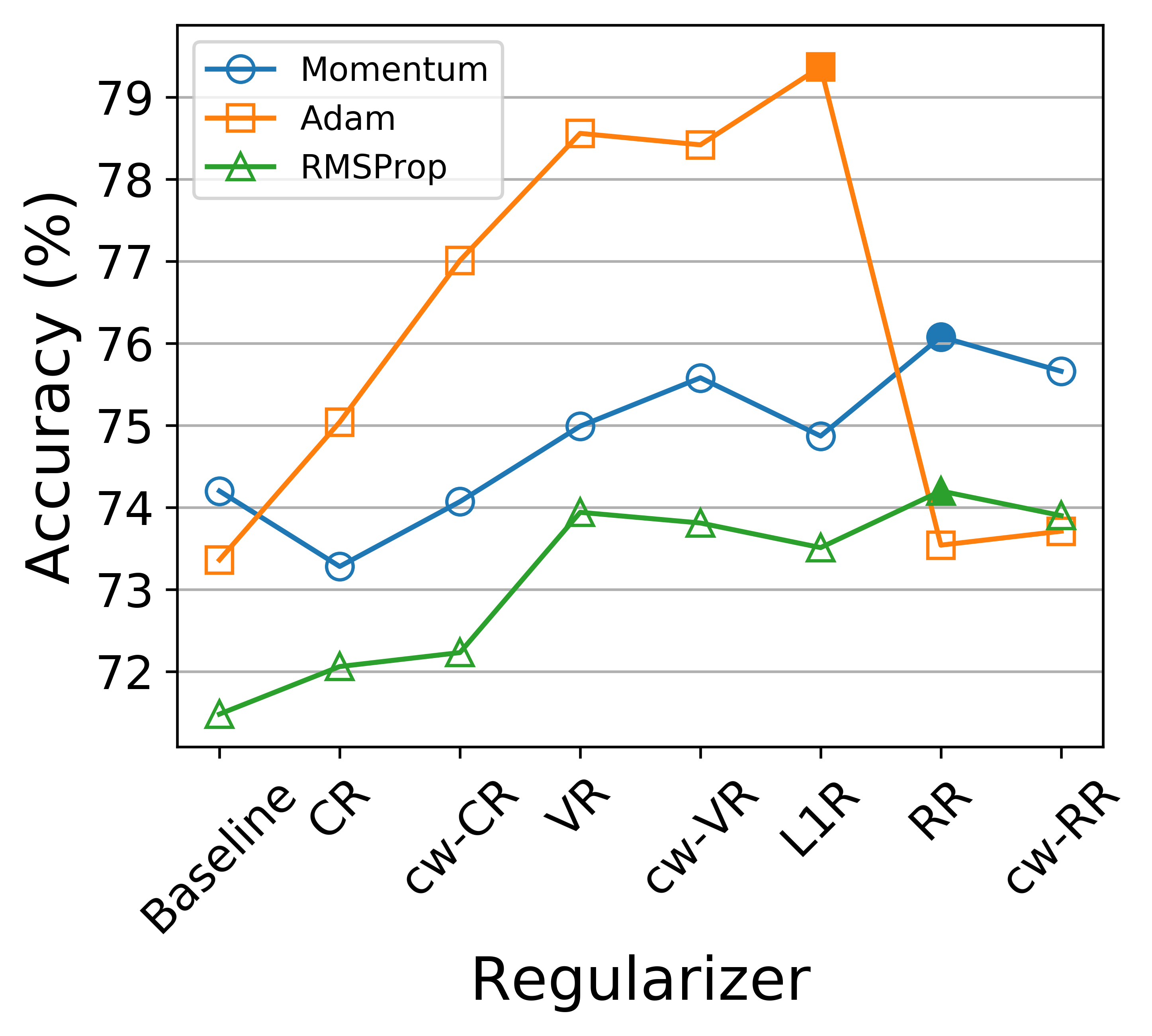}
}
\subfloat[text fot LoF][Number of classes 

                        (CNN on CIFAR-100)]{
    \includegraphics[width=0.24\textwidth]{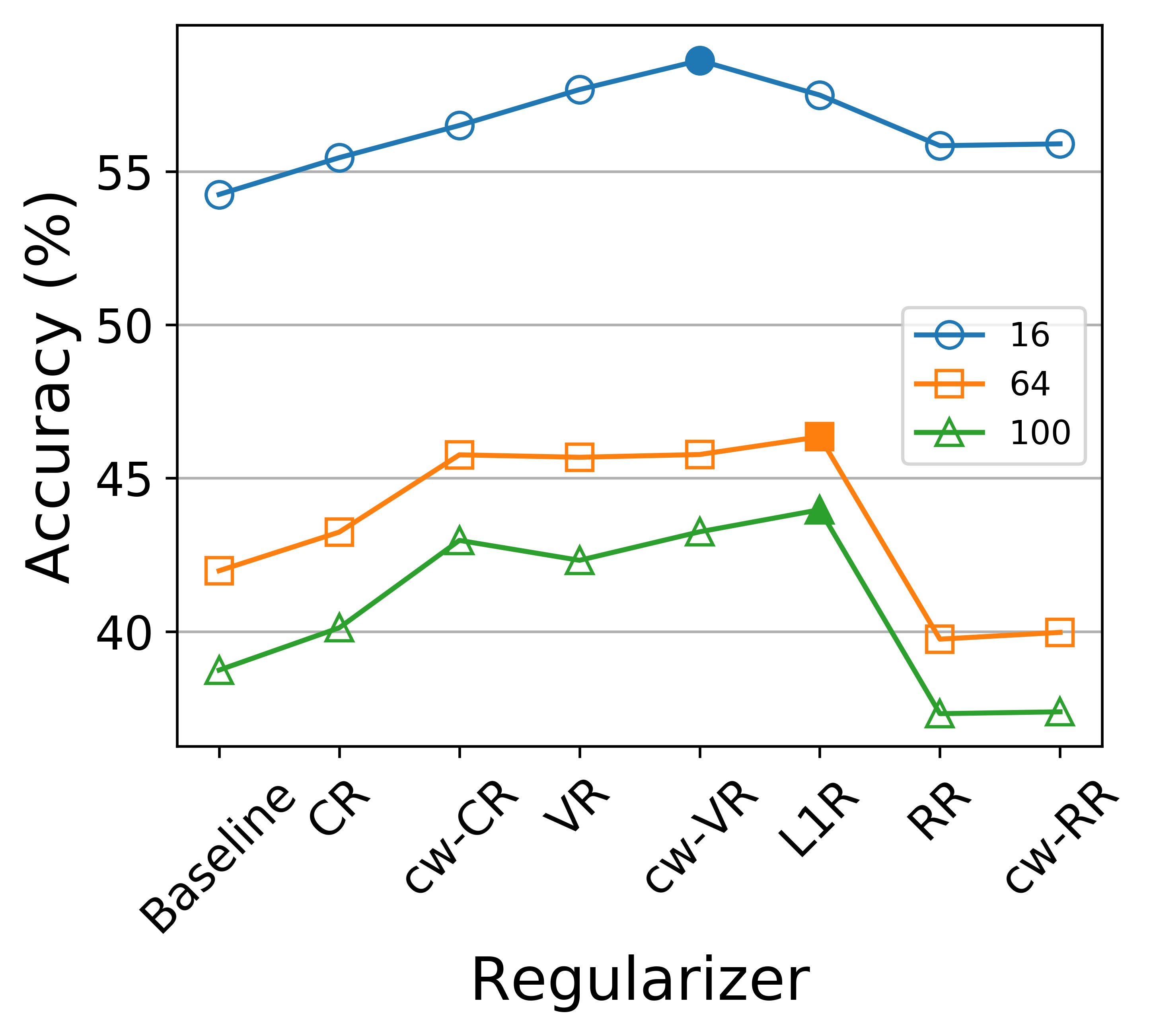}
}
\caption{Results of the `task condition' experiment. Each color indicates different task conditions such as data size, number of hidden units, choice of optimizer, and number of classes. The result shows that the seven regularizers often outperform the baseline; however, the best performing regularizer cannot be specified for the given  task condition, especially when the additional experiment results are considered together.
}
\label{fig:condition_task}
\end{figure*}

\begin{table*}[!t]
    \centering
    \caption{Test accuracy (\%) of MLP, VGG-16, and ResNet-50 models on the MNIST, the CIFAR-10/100, and the Tiny-ImageNet datasets, respectively. 
    RR and cw-RR often perform better than the others, and seven representation regularizers often mildly outperform the baseline. 
For Tiny-ImageNet, we did not experiment the class-wise regularizers because their mini-batch size is required to be much larger than the number of classes and such a configuration leads to inefficient training.}\label{tab:soph_performance}
    \vskip 0.1in
    \resizebox{1.\textwidth}{!}{%
        \begin{small}
        \begin{tabular}{lcccc}
        \toprule
            Regularizer & MLP on MNIST & VGG-16 on CIFAR-10 & VGG-16 on CIFAR-100 & ResNet-50 on Tiny-ImageNet\\ \hline
            Baseline & $97.15 \pm 0.11$ & $92.26 \pm 0.14$ & $67.11 \pm 0.44$ & $78.53 \pm 0.09$\\
            CR      & $97.50 \pm 0.05$ & $ 92.39 \pm 0.14 $ & $ 67.07 \pm 1.20 $ & $78.41 \pm 0.08$ \\
            cw-CR   & $97.51 \pm 0.10$ & $ 92.31 \pm 0.16 $ & $ 67.54 \pm 0.22 $ & - \\ 
            VR      & $97.35 \pm 0.11$ & $ 92.40 \pm 0.22 $ & $ 67.38 \pm 0.45 $ & $77.84 \pm 0.18$\\
            cw-VR   & $97.58 \pm 0.06$ & $ 92.46 \pm 0.27 $ & $ \pmb{67.63 \pm 0.32} $ & - \\ 
            L1R     & $ \pmb{97.65 \pm 0.08} $ & $92.46 \pm 0.10 $ & $ 65.56 \pm 0.31 $ & $78.54 \pm 0.13$ \\ \hline
            RR      & $97.19 \pm 0.10$ & $ 92.21 \pm 0.12 $ & $ 67.37 \pm 0.29 $ & \pmb{$78.57 \pm 0.13$} \\
            cw-RR   & $97.43 \pm 0.08$ & \pmb{$ 92.56 \pm 0.08 $} & $ 67.45 \pm 0.60 $ & - \\ 
        \bottomrule
        \end{tabular}
        \end{small}
    }
\end{table*}

\subsection{Relationship between Representation Characteristics and Performance}
To examine the relationship between representation characteristics and performance more precisely, the scatter plots of \textsc{correlation}, \textsc{sparsity}, and performance were drawn in Figure \ref{fig:relation_perf_stats}. Each circle corresponds to one characteristic and performance pair from a specific choice of model (regularizer and loss weight), and the red triangle is that of the baseline model. Only the points with comparable results to the baseline were drawn, and each model was trained and tested only once. One can observe that neither correlation nor sparsity has a clear relationship with performance. We discuss 
this phenomenon
in Section \ref{sec:discussion}.

So far, the experimental results have been shown when the regularizers are applied to the last fully-connected layer where the representation can be considered as the most processed feature set. We now investigate how differently regularizers behave when applied to different layers. In Figure \ref{fig:layer_dependency}, we apply CR (top), L1R (middle), and RR (bottom) to each layer of the 6-layer MLP. The result confirms that the regularizers perform better than the baseline (red dotted line) when applied on layers 4 and 5. Conversely, when regularizers are applied to lower layers, the performance declines, as loss weight increases even though corresponding characteristics (blue lines) can be controlled (the blue dotted lines are each layer's characteristic of the baseline model). We conjecture that this is because low-level features that should flow to the upper layers with a rich level of information can be negatively affected by putting constraints on the activations.

\subsection{Performance Improvement by Representation Regularization}

We investigate if regularizers can indeed improve the performance for a given task condition. For instance, a regularizer might be effective when the task has a small number of data examples and another regularizer might be effective when the task has a large number of classes. The following task conditions were chosen for the experiment: a learning task with 1k, 5k, or 50k data size, 32, 128, or 512 layer width, a specific dataset, a small number of classes, or a specific optimizer. Rigorously speaking, layer width and optimizer choice are not relevant to the `task', but relevant to the architecture and hyperparameter. Nonetheless, we use the term `task condition' loosely in this work. We performed experiments on the MNIST and the CIFAR-10/100 datasets using the twelve regularization setups and the four task conditions. All the regularizers are applied to the last fully-connected layer only. 


We first investigated simple MLP and CNN models. The results in Figure \ref{fig:condition_task} indicate that the regularizers are generally beneficial for improving the performance. Even though no single representation characteristic consistently outperforms the rest, it is possible to improve performance by using the regularizers as a set and by choosing the best performing regularizer for the given task. 
On the other hand, we have performed an extensive experiment in addition to the results shown in Figure \ref{fig:condition_task} (see Table \ref{table:mnist_dependency}, \ref{table:cifar10_dependency} and \ref{table:cifar100_dependency} in the supplementary materials) but we were not able to observe any meaningful relationship between a type of task condition and a type of regularizer. 

We have also investigated more sophisticated networks of VGG-16 and ResNet-50 as shown in Table \ref{tab:soph_performance}. Even for the sophisticated networks, we were able to affect representation characteristics using the regularizers and achieve a mild performance improvement. Considering that the networks have a long history of enhancement by numerous researchers and that they might have approached the best possible performance for the given task condition, even the mild improvements can be regarded as meaningful if not significant. While the performance improvements are clearly observable, again it was impossible to identify a meaningful relationship when analyzed together with the results in Table \ref{table:mnist_dependency}, \ref{table:cifar10_dependency} and \ref{table:cifar100_dependency}. 


Typically, previous works on regularizers have considered only a small number of regularizers in each work. By evaluating only a small number of regularizers over a small number of task conditions, it can be easy to identify a possibly meaningful relationship between a regularizer and a task condition. When many regularizers and many task conditions are evaluated as in our work, however, it becomes apparent that a strong relationship is extremely difficult to observe. We conclude that it is not only risky but also likely to be incorrect to imply or conjecture that a manipulation of representation can lead to an improved performance.


\section{Discussion and Future Work}
\label{sec:discussion}

\textbf{Equivalent Networks} \quad
Infinitely many global optima exist for deep neural networks \citep{du2018gradient}. By re-arranging the hidden units or by properly scaling the incoming and outgoing weights of ReLU networks, one 
can easily construct equivalent networks with different representation characteristics but with exactly same outputs \citep{dinh2017sharp}. Therefore, statistical characteristics such as correlation and covariance can be easily altered without affecting performance, simply by choosing one of the equivalent networks. The easiness of constructing equivalent networks clearly indicates that at least some of the statistical characteristics of representation do not need to have a certain property (e.g. low correlation) when the best performance is achieved. 

\textbf{Landscape of Minima} \quad
Training a deep network corresponds to finding minima of a high-dimensional non-convex loss function, and understanding the loss landscape has been an important research topic. \cite{garipov2018loss} have shown that the minima of the complex loss functions are in fact connected by simple curves over which training and test accuracy are nearly constant, and \cite{draxler2018essentially} have shown that continuous paths can be constructed between minima where the loss is essentially flat over the paths. These results can have a stronger implication than the existence of equivalent networks, because the statistical characteristics of representation over one of such connected path can be much more complicated with a wild variation.

\textbf{Learning Dynamics} \quad
In the linear least square case, a model converges along the eigenvectors of the covariance matrix at a rate depending on the magnitude of their corresponding eigenvalues \citep{lecun1991second}. Therefore, representation regularization affects not only representation characteristics but also learning dynamics. \cite{desjardins2015natural} propose a method to reparameterize the weights of the neural network by implicitly whitening each layer's activations. The method improves the learning dynamics owing to reparameterization; thus, the networks can be trained more efficiently. 
Also, \cite{combes2018learning} prove various properties of learning dynamics in deep nonlinear neural networks by studying the case of binary classification under strong assumptions such as linear separability of the data. 

While a large progress has been made in recent years, the learning dynamics of deep network still remain largely as an open problem. Together with our experiment results, it can be concluded that representation characteristics, performance, and learning dynamics are all interwoven together. While we negatively concluded on the causal and direct effect of representation characteristics to the performance, causal effects via learning dynamics is still a possibility -  representation characteristics can certainly affect learning dynamics, and learning dynamics might be able to affect the performance in a causal and explainable way. In this case, representation characteristics would indirectly affect the performance. Empirical study of all three elements remains as a possible future work.

\textbf{Generalization Bounds} \quad
\cite{jiang2019fantastic} performed a large scale study on generalization of deep networks. In the study, more than 40 complexity measures from the existing studies were chosen and investigated. The measures include traditional ones (such as VC dimension), weight matrices' norm and margin-based ones, local minima's sharpness related ones, and optimization-based ones. 
Representation characteristics, however, have been hardly considered in the theoretical and empirical studies of deep network generalization, despite representation regularizers certainly being able to affect generalization bounds through its influence on the learning of weights. The overall effect of representation characteristics can be difficult to formulate because such generalization bounds will need to depend on $p(\x)$, but perhaps a tighter bound might be obtainable for the same reason. Thus, any theoretical result might shed a light on representation characteristics' influence on the generalization performance.   

\bibliographystyle{aaai}
\bibliography{aistats}

\clearpage
\onecolumn


\aistatstitle{Supplementary Materials}

\section*{A.\quad Class-wise Statistics and Representation Regularizers}
Based on the notations in Section \ref{sec:characteristics}, we define class-wise statistics that are calculated using only samples of class $k$, out of a total of $K$ labels in the mini-batch. Class-wise mean, covariance, and variance are defined as follows. 
\begin{align}
    \mu_{z_{l,i}}^{(k)} &= \EE_{n \in S_k}[z_{l,i}^{n}]. \label{eq:mean_cw} \\
    \textit{c}_{i,j}^{(k)} &= \EE_{n \in S_k}[(z_{l,i}^{n} - \mu_{z_{l,i}}^{(k)})(z_{l,j}^{n} - \mu_{z_{l,j}}^{(k)})]. \label{eq:covariance_cw}  \\  
    \textit{v}_{z_{l,i}}^{(k)} &= \textit{c}_{i,i}^{(k)}.   \label{eq:variance_cw}
\end{align}
Here, $S_k$ is the set that contains the indices of the samples with the class label $k$. Note that superscripts with and without parenthesis indicate class label and sample index, respectively. 

Penalty loss functions of the representation regularizers are summarized in Table \ref{table:loss_function}.
\begin{table*}[!ht]
\caption{Penalty loss functions of representation regularizers.}
\begin{center}
\vskip -0.15in
\begin{tabular}{rll}
\toprule
\multicolumn{1}{c}{Symbol} & \multicolumn{1}{c}{Penalty loss function}  & \multicolumn{1}{c}{Description of regularization term}  \\ \hline
$\displaystyle{\Omega}_{CR}$ & $=\sum_{i\neq j} (c_{i,j})^{2} $    &  \textit{Covariance} of representations calculated from all-class samples. \\ \hline
$\displaystyle{\Omega}_{cw{\text -}CR}$ & $=\sum_k \sum_{i\neq j} (c_{i,j}^{(k)})^{2}$   & \textit{Covariance} of representations calculated from \textbf{the same class samples}.\\ \hline
$\displaystyle{\Omega}_{VR}$ & $=\sum_i v_{i}$                                               & \textit{Variance} of representations calculated from all-class samples.  \\ \hline
$\displaystyle{\Omega}_{cw{\text -}VR}$ & $=\sum_k \sum_i v_{z_{l,i}}^{(k)}$              & \textit{Variance} of representations calculated from \textbf{the same class samples}.  \\ \hline 
$\displaystyle{\Omega}_{L1R}$ & $=\sum_n \sum_i |z_{l,i}^{n}|$    &  \textit{Absolute amplitude}  of representations calculated from all-class samples.\\ \hline
$\displaystyle\Omega_{RR}$ & $=\norm{\Z_l}_{F}^{2}/\norm{\Z_l}_{2}^{2}$ & \textit{Stable rank} of representations calculated from all-class samples. \\ \hline
$\displaystyle\Omega_{cw{\text -}RR}$ & $=\sum_k(\norm{\Z_l^{(k)}}_{F}^{2}/\norm{\Z_l^{(k)}}_{2}^{2})$ & \textit{Stable rank} of representations calculated from \textbf{the same class samples}.\\ 
\bottomrule
\end{tabular}
\label{table:loss_function}
\end{center}
\vskip -0.2in
\end{table*}


\section*{B.\quad Experiment Details}
\subsection*{B.1\quad Parameters of CNNs}

A 6-layer MLP with 100 units per hidden layer was used for MNIST image classification tasks. A CNN with four convolutional layers and one fully-connected layer was used for CIFAR-10/100 image classification tasks. Architecture details are described in Table \ref{table:hyperparameters}.

\begin{table*}[!ht]
\centering
\caption{Architecture hyperparameters of CIFAR-10/100 CNN model.}
\begin{tabular}{cc}
\hline
Layer                 & Parameter  \\ \hline
Convolutional layer-1 & Number of filters=32, Filter size=3 $\times$ 3, Convolution stride=1 \\ \hline    
Convolutional layer-2 & Number of filters=64, Filter size=3 $\times$ 3, Convolution stride=1 \\  \hline   
Max-pooling layer-1   & Pooling size=2 $\times$ 2, Pooling stride=2              \\ \hline
Convolutional layer-3 & Number of filters=128, Filter size=3 $\times$ 3, Convolution stride=1 \\ \hline     
Max-pooling layer-2   & Pooling size=2 $\times$ 2, Pooling stride=2              \\ \hline
Convolutional layer-4 & Number of filters=128, Filter size=3 $\times$ 3, Convolution stride=1 \\ \hline    
Max-pooling layer-3   & Pooling size=2 $\times$ 2, Pooling stride=2              \\ \hline
Fully connected layer & Number of units=128             \\ \hline             
\end{tabular}
\label{table:hyperparameters}
\end{table*}

\subsection*{B.2\quad Parameters of Sophisticated Networks}

\textbf{VGG-16}\hspace{0.1cm} We adapt the state-of-the-art VGG-16 network for the CIFAR-10 and CIFAR-100 datasets from (https://github.com/geifmany/cifar-vgg). To manipulate the properties of the representation statistics, for the penultimate fully-connected layer, we exclude additional processes (e.g. batch normalization, dropout), then apply the representation regularizer. All hyperparameters are chosen to be the same as the SOTA network. While fine-tuning the network, the network was trained for only 100 epochs with a smaller initial learning rate of 0.001.

\textbf{Resnet}\hspace{0.1cm} We adapt Resnet-50 for Tiny-Imagenet and Resnet-18 for Imagenet. To manipulate the properties of the representation characteristics, we add a fully-connected layer following the last average pooling layer. Then we apply the representation regularizers to the added layer. All hyperparameters are chosen to be the same as the SOTA network. While fine-tuning the network, the network was trained for only 100 epochs with smaller initial learning rates of 0.02 (for Tiny-Imagenet) and 0.0001 (for Imagenet).

\section*{C.\quad Principal Component Analysis of Learned Representations}

\begin{figure*}[!bt]
  \centering
  \includegraphics[width=0.95\textwidth]{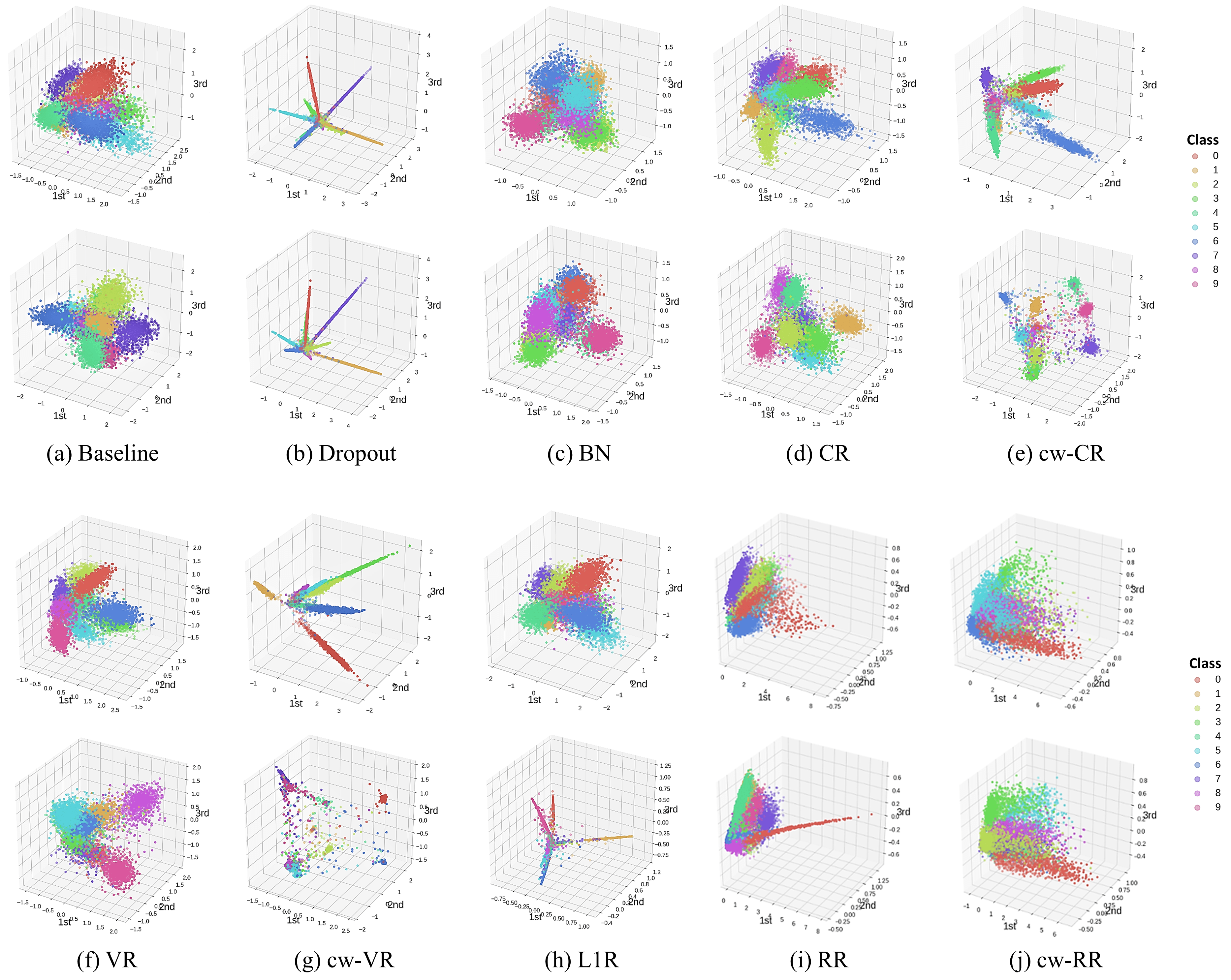}
  \caption{Top three principal components of learned representations. For each regularization, the upper one shows the scatter plot of activations before passing through the ReLU, and the lower one shows the scatter plot of activations after passing through the ReLU. Note that the top three PCA directions are affected when ReLU converts the negative values to zero, and thus the upper and lower plots look different.}
  \label{fig:pca}
  \vskip -0.3in
\end{figure*}

In Figure \ref{fig:scatter}, we have shown activation histogram of a single unit and activation scatter plots of two randomly chosen units. The plots clearly demonstrate the variations in representation characteristics as different representation regularizers are applied. Alternatively, we can choose the directions with the largest variations using PCA, and generate similar plots. In Figure \ref{fig:pca}, the top three PCA directions were chosen to generate activation scatter plots. As in Figure \ref{fig:scatter}, variations in representation characteristics can be conspicuously observed.

\clearpage
\section*{D.\quad Result of Condition Tasks}
\subsection*{D.1\quad Task Conditions}
Experimental conditions are listed as follows. (Default conditions are shown in bold.) 

\noindent $\bullet$ Training data size: 1k, 5k, \textbf{50k} \\
\noindent $\bullet$ Layer width: (MNIST) 2, 8, \textbf{100} / (CIFAR-10/100): 32, \textbf{128}, 512 \\
\noindent $\bullet$ Optimizer (CIFAR-10): \textbf{Adam (lr=0.0001)}, Momentum (lr=0.01, momentum=0.9), RMSProp (lr=0.0001) \\
\noindent $\bullet$ Number of classes (CIFAR-100): 16, 64, \textbf{100}  
\subsection*{D.2\quad Result of Condition Tasks}

\begin{table*}[ht]
\caption{Condition experiment results for the MNIST MLP model. A 6-layer MLP with 100 units per hidden layer was used.
The best performing regularizer in each condition (each column) is shown in bold; 
other regularizers whose performance range overlaps with the best one are highlighted in green. For the default condition, the standard values of data size=50k and layer width=100 were used, 
and the Adam optimizer was applied. For other columns, all the conditions were the same as the default,
except for the condition indicated on the top part of the columns.}
\vskip -0.2in
\begin{center}
\begin{small}
\newcommand*{\bgcblack}[1]{\cellcolor{green!20}\pmb{#1}}
\newcommand*{\bgcbb}[1]{\cellcolor{green!20}{#1}}
\begin{tabular}{lcccccr}
\hline
\multirow{2}{*}{Regularizer} & \multirow{2}{*}{Default} & \multicolumn{2}{c}{Data Size}      & \multicolumn{2}{c}{Layer Width}     \\ \cmidrule{3-6} 
                             &                          & 1k               & 5k              & 2                & 8                \\ \hline
Baseline                     & $97.15 \pm 0.11$          & $88.59 \pm 0.19$ & $94.00 \pm 0.07$ & $68.38 \pm 0.07$ & $89.48 \pm 0.57$ \\ \hline
L1W                          & $97.15 \pm 0.06$          & $88.36 \pm 0.27$ & $94.96 \pm 0.11$ & $68.33 \pm 0.15$ & $88.98 \pm 0.58$ \\ 
L2W                          & $96.98 \pm 0.40$          & $88.62 \pm 0.18$ & $94.14 \pm 0.10$ & $68.34 \pm 0.13$ & $89.35 \pm 0.23$ \\ \hline
Dropout                      & $97.30 \pm 0.08$          & \bgcblack{$89.71 \pm 0.23$} & \bgcblack{$94.41 \pm 0.11$} & $37.91 \pm 1.32$ & $86.06 \pm 1.05$ \\ 
BN                           & $97.19 \pm 0.12$          & $89.19 \pm 0.04$ & \bgcbb{$94.40 \pm 0.10$} & $57.92 \pm 0.93$ & \bgcblack{$92.49 \pm 0.58$}  \\ \hline
CR                           & $97.50 \pm 0.05$          & $88.37 \pm 0.24$ & $93.95 \pm 0.06$ & $65.20 \pm 0.25$ & $89.75 \pm 0.74$ \\ 
cw-CR                        & \bgcbb{$97.51 \pm 0.10$}          & $89.38 \pm 0.05$ & \bgcbb{$94.20 \pm 0.15$} & $68.50 \pm 0.11$ & $89.19 \pm 1.11$ \\ 
VR                           & $97.35 \pm 0.11$          & $85.58 \pm 0.14$ & $93.10 \pm 0.22$ & $67.61 \pm 0.13$ & $90.78 \pm 0.28$  \\ 
cw-VR                        & \bgcbb{$97.58 \pm 0.06$}          & \bgcbb{$89.56 \pm 0.18$} & $94.10 \pm 0.12$ & \bgcblack{$69.66 \pm 0.06$} & $89.99 \pm 0.63$ \\
L1R                          & \bgcblack{$97.65 \pm 0.08$}          & $88.40 \pm 0.20$ & $93.80 \pm 0.13$ & $35.61 \pm 0.26$ & $11.35 \pm 0.00$ \\ 
RR                           & $97.19 \pm 0.10$          & $89.08 \pm 0.17$ & $93.39 \pm 0.05$ & $61.65 \pm 0.20$ & $87.69 \pm 0.16$  \\ 
cw-RR                        & $97.43 \pm 0.08$          & $89.11 \pm 0.19$ & $93.40 \pm 0.17$ & $61.43 \pm 0.12$ & $87.37 \pm 0.39$ \\ \hline
\end{tabular}
\label{table:mnist_dependency}
\end{small}
\end{center}
\end{table*}

\begin{table*}[!ht]
\newcommand*{\bgcblack}[1]{\cellcolor{green!20}\pmb{#1}}
\newcommand*{\bgcbb}[1]{\cellcolor{green!20}{#1}}
\caption{Condition experiment results for the CIFAR-10 CNN model (Test error \%).
The best performing regularizer in each condition (each column) is shown in bold; other regularizers whose performance range overlaps with the best one are highlighted in green. 
For the default condition, the standard values of data size=50k, layer width=128,
and the Adam optimizer was applied. For the others, all the conditions were the same as the default, 
except for the condition indicated on the top part of the columns. Regularizers were applied to the fully-connected layer.}\label{tab:performance}
\centering
\begin{small}
\resizebox{\textwidth}{!}{%
\begin{tabular}{lccccccc}
\hline
\multirow{2}{*}{Regularizer} & \multirow{2}{*}{Default} & \multicolumn{2}{c}{Data Size}       & \multicolumn{2}{c}{Layer Width} & \multicolumn{2}{c}{Optimizer}    \\ \cmidrule{3-8}
                             &                          & 1k               & 5k               & 32                & 512         & {Momentum} & {RMSProp}       \\ \hline
Baseline                     & $73.36 \pm 0.16$         & $43.93 \pm 0.36$ & $56.05 \pm 0.43$ & $71.46 \pm 0.63$ & $71.48 \pm 1.06$             & $74.22 \pm 0.37$ & $71.48 \pm 1.21$ \\ \hline
L1W                          & $73.54 \pm 0.39$         & $43.36 \pm 0.91$ & $55.68 \pm 0.66$ & $71.35 \pm 1.14$ & $72.04 \pm 0.72$             & $74.27 \pm 0.40$ & $71.70 \pm 0.99$ \\
L2W                          & $74.29 \pm 0.98$            & $43.43 \pm 0.22$ & $55.13 \pm 0.81$ & $71.46 \pm 0.30$  & $72.21 \pm 0.83$    & $73.65 \pm 0.54$ & $71.98 \pm 0.88$ \\ \hline
Dropout                      & $73.63 \pm 0.21$         & $43.89 \pm 0.83$ & $55.22 \pm 0.41$ & $72.34 \pm 0.51$ & $71.57 \pm 0.88$             & $74.05 \pm 0.57$ & $72.31 \pm 0.38$ \\
BN                           & $68.03 \pm 3.10$         & $43.51 \pm 0.32$ & $56.25 \pm 0.76$ & $71.17 \pm 0.47$ & $71.80 \pm 0.40$             & $74.50 \pm 0.55$ & $71.62 \pm 0.86$ \\ \hline
CR                           & $75.04 \pm 0.63$         & $42.60 \pm 2.11$  & $54.84 \pm 0.94$ & $73.55 \pm 0.22$ & $71.35 \pm 1.21$            & $73.28 \pm 0.61$   & $72.06 \pm 0.43$  \\
cw-CR                        & $77.01 \pm 0.58$         & \bgcbb{$46.50 \pm 1.05$}  & $57.85 \pm 0.64$ & $73.60 \pm 0.62$  & $71.46 \pm 1.01$    & $74.07 \pm 0.59$    & $72.23 \pm 0.88$  \\ 
VR                           & \bgcbb{$78.56 \pm 0.88$}         & \bgcbb{$46.10 \pm 0.97$}  & $57.67 \pm 0.57$ & \bgcblack{$75.04 \pm 0.26$} & \bgcbb{$73.39 \pm 0.47$} & $74.99 \pm 0.41$    & $73.94 \pm 0.72$  \\
cw-VR                        & $78.42 \pm 0.21$         & \bgcblack{$48.07 \pm 1.09$} & $57.00 \pm 0.95$    & \bgcbb{$74.19 \pm 0.64$} & \bgcblack{$73.54 \pm 0.25$}     &  \bgcbb{$75.58 \pm 0.31$}   & \bgcbb{$73.81 \pm 1.35$}  \\ 
L1R                          & \bgcblack{$79.37 \pm 0.50$}         & \bgcbb{$47.61 \pm 0.99$} & \bgcblack{$59.08 \pm 0.33$} & \bgcbb{$74.51 \pm 0.61$} & $72.19 \pm 0.43$     & $74.87 \pm 0.52$  & \bgcbb{$73.51 \pm 0.96$}\\ 
RR                        & $73.54 \pm 0.25$         & $42.91 \pm 1.08$ & $55.65 \pm 1.09$ & $73.42 \pm 0.66$ & \bgcbb{$73.13 \pm 0.58$}  &    \bgcblack{$76.08 \pm 0.37$}  &  \bgcblack{$74.20 \pm 0.85$} \\
cw-RR                        & $73.71 \pm 0.41$         &  $42.45 \pm 0.46$ & $55.29 \pm 1.59$  & $73.38 \pm 0.77$ & \bgcbb{$72.88 \pm 0.46$}    &    \bgcbb{$75.66 \pm 0.27$} & \bgcbb{$73.90 \pm 0.59$}  \\ \hline
\end{tabular}%
}
\end{small}
\vskip -0.1in
\label{table:cifar10_dependency}
\end{table*}

\begin{table*}[ht]
\caption{Condition experiment results for the CIFAR-100 CNN model. 
The best performing regularizer in each condition (each column) is shown in bold, 
and other regularizers whose performance range overlaps with the best one are highlighted in green. For the default condition, the standard values of data size=50k, layer width=128, 
and number of classes=100 were used.
For the other columns, all the conditions were the same as the default, 
except for the condition indicated on the top part of the columns.
Regularizers were applied to the fully-connected layer.}
\vskip -0.2in
\begin{center}
\newcommand*{\bgcblack}[1]{\cellcolor{green!20}\pmb{#1}}
\newcommand*{\bgcbb}[1]{\cellcolor{green!20}{#1}}
\resizebox{\textwidth}{!}{%
\begin{small}
\begin{tabular}{lccccccc}
\hline
\multirow{2}{*}{Reg.} & \multirow{2}{*}{Default}                     & \multicolumn{2}{c}{Data Size}                                                               & \multicolumn{2}{c}{Layer Width}                                                             & \multicolumn{2}{c}{Number of Classes}       \\ \cmidrule{3-8}                                                
                             &                                              & 1k                                           & 5k                                           & 32                                           & 512                                          & 16                                           & 64                                           \\ \hline
Baseline                     & $38.74 \pm 0.52$                             & \bgcbb{$9.11 \pm 0.30$}    & \bgcbb{$17.79 \pm 0.72$}    & $37.59 \pm 0.34$                             & $38.70 \pm 0.64$                             & $54.25 \pm 0.73$                             & $41.98 \pm 0.40$                             \\ \hline
L1W                          & $39.03 \pm 0.64$                             & $8.67 \pm 0.37$                             & $17.70 \pm 0.60$                               & $37.77 \pm 0.58$                             & $39.08 \pm 0.47$                             & $54.92 \pm 1.53$                             & $41.92 \pm 1.18$                             \\
L2W                          & $39.77 \pm 0.31$                             & \bgcbb{$9.47 \pm 0.39$}    & \bgcbb{$17.95 \pm 0.70$}    & $37.22 \pm 0.36$                             & $38.45 \pm 0.99$                             & $54.72 \pm 1.59$                             & $42.53 \pm 0.66$                             \\ \hline
Dropout                      & $36.12 \pm 0.72$                             & \bgcblack{$9.78 \pm 0.48$} & \bgcbb{$18.32 \pm 0.81$}    & $35.92 \pm 0.99$                             & $35.69 \pm 0.37$                             & $54.27 \pm 1.57$                             & $40.86 \pm 0.46$                             \\
BN                           & $39.07 \pm 0.39$                             & $8.82 \pm 0.36$                             & \bgcbb{$17.99 \pm 0.58$}    & \bgcbb{$37.82 \pm 1.49$}    & $37.84 \pm 0.57$                             & $55.45 \pm 1.43$                             & $42.28 \pm 0.66$                             \\ \hline
CR                           & $40.12 \pm 0.50$                             & $8.30 \pm 0.14$                             & $17.53 \pm 0.41$                             & \bgcblack{$39.53 \pm 0.63$} & $39.30 \pm 0.94$                             & $55.45 \pm 1.10$                             & $43.24 \pm 0.86$                             \\
cw-CR                        & \bgcbb{$42.97 \pm 0.73$}    & \bgcbb{$9.15 \pm 0.29$}    & \bgcbb{$18.71 \pm 0.62$}    & \bgcbb{$38.59 \pm 0.67$}    & $41.98 \pm 0.25$                             & $56.50 \pm 1.21$                             & \bgcbb{$45.76 \pm 0.64$}    \\
VR                           & \bgcbb{$42.32 \pm 0.94$}    & $8.57 \pm 0.32$                             & \bgcbb{$18.15 \pm 0.38$}    & \bgcbb{$38.65 \pm 0.45$}    & \bgcbb{$43.13 \pm 0.74$}    & \bgcbb{$57.67 \pm 1.03$}    & \bgcbb{$45.68 \pm 0.40$}    \\
cw-VR                        & \bgcbb{$43.25 \pm 0.64$}    & \bgcbb{$9.55 \pm 0.22$}    & \bgcblack{$18.97 \pm 0.57$} & \bgcbb{$39.33 \pm 0.59$}    & \bgcbb{$43.09 \pm 0.73$}    & \bgcblack{$58.62 \pm 0.53$} & \bgcbb{$45.77 \pm 1.06$}    \\
L1R                          & \bgcblack{$43.97 \pm 0.81$} & $8.85 \pm 0.35$                             & \bgcbb{$18.02 \pm 0.47$}    & \bgcbb{$38.89 \pm 0.31$}    & \bgcblack{$43.54 \pm 0.62$} & \bgcbb{$57.49 \pm 1.43$}    & \bgcblack{$46.35 \pm 1.00$} \\ 
RR                           &   $37.32 \pm 0.35$                                           & $8.80 \pm 0.27$                             & \bgcbb{$18.68 \pm 0.36$}    & $31.46 \pm 0.46$    & $40.71 \pm 0.32$    & $55.84 \pm 0.80$    & $39.75 \pm 0.35$    \\
cw-RR                        &      $37.38 \pm 0.31$                                        & \bgcbb{$9.38 \pm 0.34$}    & \bgcbb{$18.43 \pm 0.14$} & $31.89 \pm 0.31$    & $40.75 \pm 0.29$   & $55.90 \pm 0.65$ & $39.97 \pm 0.41$    \\ \hline
\end{tabular}%
\end{small}
}
\label{table:cifar100_dependency}
\end{center}
\end{table*}




\end{document}